# ICED 2020 proceedings:
# Artificial Intelligence enabled Smart Learning

**Debdeep Bose[1], Pathan Faisal Khan[2]**

*Department of Computing, Letterkenny Institute of Technology*
*Port Road, Gortlee, Letterkenny*
*Co. Donegal, Ireland*

*Abstract*

Artificial Intelligence (AI) is a discipline of computer science that deals with machine intelligence. It is essential to bring AI into the context of learning because it helps in analysing the enormous amounts of data that is collected from individual students, teachers and academic staff. The major priorities of implementing AI in education are making innovative use of existing digital technologies for learning, and teaching practices that significantly improve traditional educational methods.

The main problem with traditional learning is that it cannot be suited to every student in class. Some students may grasp the concepts well, while some may have difficulties in understanding them and some may be more auditory or visual learners. The World Bank report on education has indicated that the learning gap created by this problem causes many students to drop out (*World Development Report*, 2018). Personalised learning has been able to solve this grave problem. Brainly is one example: an AI-based knowledge-sharing social network where students post and respond to questions asked by other students. This collaboration, along with personalisation-based machine learning algorithms for networking features, has made Brainly a successful platform with 8,000 items responded to every hour. AI is also used in the classroom: China's largest AI education platform "Squirrel AI"[3] has successfully implemented its system in many cities to provide personalised learning. They claim that their system was better at improving math test scores than experienced teachers teaching in a four-day experiment program conducted in October 2017 (Dickson, 2017). These are just a couple of examples of the use of AI inside the classroom to enhance learning activity. A few more, such as "Osmo"[4] and "Classcraft"[5], are integrated into K-12 programmes. It has also been predicted that personalised teaching methodologies using AI will reduce the cost of education, which is quite high even in developed countries. Master Learner is an AI-powered education platform from Shanghai, and they have claimed that "we can make our training fee as cheap as drinking a Coke every day for a year" (*China Turns to Artificial Intelligence to Boost Its Education System*, 2017).

## 1    Introduction

Smart Learning includes new educational contexts in which the focus is students' use of the technology at their fingertips. This depends not only on the software and hardware available but on how they are used in the classes. The main problem with traditional learning is that it cannot be suited to every child. The World Bank report on education has indicated that the learning gap created by disparities in this regard causes many students to drop out of school. Personalised learning has been able to solve this grave problem. The theory and development of computer systems able to perform tasks which typically require human intelligence is known as Artificial Intelligence (AI), and AI comes into play here; see the examples below.

---

[1] debdeep.bose99@gmail.com
[2] faisal3325@gmail.com
[3] http://squirrelai.com
[4] https://www.playosmo.com
[5] https://www.classcraft.com

## 2   Classcraft

### 2.1   The system

Launched in 2015, Classcraft is an award-winning, teacher-friendly gamification tool that is now used in more than 50,000 classrooms in 75 countries in 11 languages. The Quebec- and New York City-based education technology company uses gaming principles to address student motivation. It utilises gaming principles to foster social-emotional development and personalised learning, enabling educators to adapt curricula and teaching style to an individual student's needs. Highly customisable, this revolutionary educational approach can be adapted to any subject and has been proven to be very effective at improving student motivation, increasing student engagement and creating a positive classroom community by fostering secure team building.

### 2.2   Facts

1. Nearly 54% of the EU population plays video games, which means that there are some 250 million players in the EU. Roughly half of players are female (46%), and 54% are male (Games in Society, 2019).
2. 58% of parents play video games with their kids to socialise with them (Hallifax et al., 2019).
3. A study funded by the British Academy and published in the journal Computers in Human Behaviour, found that 13-to-14-year-old girls classed as "massive gamers" – those playing over nine hours a week – were three times more likely to pursue a PSTEM (physical science, technology, engineering, and math) degree compared to girls who were non-gamers ("Geek Girl" Gamers Are More Likely to Study Science and Technology Degrees | University of Surrey, 2019).
4. A study by Centre de Liaison sure Intervention et la Prévention Psychosociales, or (CLIPP) showed that gamification could help prevent bullying (Classcraft - Gamification in Education, 2019).

## 3   Alta

### 3.1   The system

Alta is a software product built on a personalised learning engine for students pursuing higher education by Knewton, a New-York-based adaptive learning company. Knewton released Alta in January 2018 after ten years of experience with publishers. It is powered by their in-house high-quality content curated with long experience in the industry. The software is available to students as a standalone package and to universities and institutes as a comprehensive tool for all students. The product covers courses in mathematics, chemistry, economics and statistics. Courses contain textual, graphical and video-based content. The software is available for students as a standalone version for just $39.95. Alta's mobile app has made it a mobile software for every student to use on the go. The best thing about Alta is that students are not just left alone with the software; there is 24/7 online chat support for student doubts and queries.

### 3.2   Facts

1. 87% of the time, students using Alta completed their assignments with proficiency.
2. Of students who struggled in their assignments, 82% of them completed them.

3. Students who used Alta and did not complete an assignment scored 55%, while those who completed their task scored an average of 81%.
4. Students who were struggling have shown improvement in their scores, from being at 40% to achieving 78%.
5. 85% of students feel that Alta is improving their skills.
6. Arizona State University has claimed that after implementing Knewton's adaptive learning technology, on which Alta is based, there has been a decrease in dropout rates from 13% to 6% and a rise in pass rates from 66% to 75%.

## 4   Squirrel AI learning

### 4.1   The system

Squirrel AI is a Shanghai-based after-school tutoring company with nearly 2000 physical classrooms in China. Squirrel AI was founded in 2014 by Derek Li Haoyang after he stepped down as CEO from his previous education company, which featured an IPO. Using a laptop computer with the company's software installed on it, students study their lessons in a classroom supervised by a teacher of the respective subject.

Squirrel AI's main motive was to address the problems faced in the education system: lack of personalised attention in classrooms and unequal distribution of educational opportunities. The inefficient, rigid education system has decreased students' enthusiasm for learning; this motivated Derek Li to build what is China's most extensive AI-powered education product. Squirrel AI's scope and reach are impressive. However, the concept behind adaptive learning systems like Squirrel AI and others will not make teaching professionals obsolete any time soon. Instead, Squirrel AI is designed to support and augment the work of teachers by taking away the need to teach the "nuts and bolts" of each course (*Building Personalized Education With AI Adaptive Learning - AI Business*, 2019).

### 4.2   Facts

1. It has been shown that the Squirrel AI system can teach 48 knowledge points in eight hours on average, whereas a human teacher can explain 28 knowledge points in the same period.

## 5   Conclusions

In summary, Artificial Intelligence enabled Smart Learning is the next logical phase in the introduction of technology to classrooms and educational centres. The global educational landscape has been changing with the introduction to new state-of-the-art intelligent environments; a few have termed this "climate change" in education. Our paper (originally a poster) presents some of the latest successful software and its implementation, all based on AI enabled Smart Learning. However, this is not just about selecting a tool or technology – even though technology (like it or not) is essential to learning and is a primary part of every industry and because it changes so quickly students are better off learning about it sooner. More than this, it is essential to deploy a proven methodology that works with students and develops their skills in a progressive, natural and effective way.

We thus bear in mind that the technological and pedagogical advancements described above are not meant to supersede current teaching and learning education systems, but rather to provide a holistic spectrum of complementary supporting tools which harness and exploit this emerging paradigm to its full potential for smarter education (*Smart Learning for the Next Generation Education Environment*, 2014).

*References*

*AI Business* (2019). *Building Personalized Education with AI Adaptive Learning.* https://aibusiness.com/document.asp?doc_id=760699&site=aibusiness

*China turns to artificial intelligence to boost its education system* (2017, October 14). South China Morning Post. https://www.scmp.com/tech/science-research/article/2115271/china-wants-bring-artificial-intelligence-its-classrooms-boost

*Gamification in Education* (2019). Classcraft. https://www.classcraft.com/gamification/

Dickson, B. (2017). *How Artificial Intelligence enhances education*. The Next Web. https://thenextweb.com/artificial-intelligence/2017/03/13/how-artificial-intelligence-enhances-education/

*Games in Society* (2019). ISFE. https://www.isfe.eu/games-in-society/

University of Surrey (2019). *"Geek Girl" gamers are more likely to study science and technology degrees.* https://www.surrey.ac.uk/news/geek-girl-gamers-are-more-likely-study-science-and-technology-degrees

Hallifax, S., Serna, A., Marty, J.-C., & Lavoué, É. (2019). Adaptive Gamification in Education: A Literature Review of Current Trends and Developments. In M. Scheffel, J. Broisin, V. Pammer-Schindler, A. Ioannou, & J. Schneider (Eds.), *Transforming Learning with Meaningful Technologies* (pp. 294–307). Springer International Publishing. https://doi.org/10.1007/978-3-030-29736-7_22

*Research Behind ALEKS - Knowledge Space Theory* (2019). https://www.aleks.com/about_aleks/knowledge_space_theory

*Smart Learning for the Next Generation Education Environment* (2014). ResearchGate. https://www.researchgate.net/publication/286679937_Smart_Learning_for_the_Next_Generation_Education_Environment

*World Development Report 2018: Learning to Realize Education's Promise*. (2018). [Text/HTML]. World Bank. https://www.worldbank.org/en/publication/wdr2018